\documentclass{article}

\usepackage[preprint]{neurips_2024}

\usepackage[utf8]{inputenc} 
\usepackage[T1]{fontenc}    
\usepackage{hyperref}       
\usepackage{url}            
\usepackage{booktabs}       
\usepackage{amsfonts}       
\usepackage{nicefrac}       
\usepackage{microtype}      
\usepackage{xcolor}         
\AtBeginDocument{%
  }

\usepackage{float}
\usepackage[countmax]{subfloat}
\usepackage{subcaption}
\usepackage{amsmath,amsfonts,amsmath}
\usepackage{listings}
\usepackage{algorithmic}
\usepackage{graphicx}
\usepackage{color}
\usepackage{xcolor}
\usepackage{textcomp}
\usepackage{xurl}

\usepackage{booktabs}
\usepackage{nicefrac}
\usepackage{lipsum}
\usepackage{multicol}
\usepackage{multirow}
\usepackage{makecell}
\usepackage{tabularx}
\usepackage{cleveref}
\usepackage[separate-uncertainty=true,group-separator={,}]{siunitx}
\usepackage{datetime}
\usepackage[subtle]{savetrees}
\usepackage{microtype}
\usepackage{comment}
\usepackage{todonotes}
\usepackage{commands}
\usepackage[inline]{enumitem}
\usepackage{acronym}
\usepackage{hyperref}
\usepackage{uppaal}

\acrodef{smc}[SMC]{Statistical Model Checking}
\acrodef{nlp}[NLP]{Natural Language Processing}
\acrodef{lm}[LM]{Language Model}
\acrodefplural{lm}[LMs]{Language Models}
\acrodef{llm}[LLM]{Large Language Model}
\acrodefplural{llm}[LLMs]{Large Language Models}
\acrodef{sha}[SHA]{Stochastic Hybrid Automaton}
\acrodefplural{sha}[SHA]{Stochastic Hybrid Automata}
\acrodef{mitl}[MITL]{Metric Interval Temporal Logic}
\acrodef{sleec}[SLEEC]{social, legal, ethical, empathetic, and cultural}
\acrodef{ode}[ODE]{Ordinary Differential Equation}
\acrodefplural{ode}[ODEs]{Ordinary Differential Equations}

\graphicspath{{}}

\title{``Do You Understand How I Feel?'': \\ Towards Verified Empathy in Therapy Chatbots}

\author{%
  Francesco Dettori\thanks{These authors contributed equally to this work.} \\
  Université Paris-Saclay, \\
  Centre National de la Recherche Scientifique \\
  Paris, France \\
  \texttt{francesco.dettori@universite-paris-saclay.fr} \\
  \href{https://orcid.org/0009-0008-8482-7502}{ORCID: 0009-0008-8482-7502} \\
  \And
  Matteo Forasassi\footnotemark[1] \\
  TU Wien \\
  Vienna, Austria \\
  \texttt{matteo.forasassi@tuwien.ac.at} \\
  \href{https://orcid.org/0009-0000-1073-7151}{ORCID: 0009-0000-1073-7151} \\
  \And
  Lorenzo Veronese\footnotemark[1] \\
  Politecnico di Milano \\
  Milan, Italy \\
  \texttt{lorenzo.veronese@polimi.it} \\
  \href{https://orcid.org/0000-0001-9204-3363}{ORCID: 0000-0001-9204-3363} \\
  \AND
  Livia Lestingi\thanks{Corresponding author.} \\
  Politecnico di Milano \\
  Milan, Italy \\
  \texttt{livia.lestingi@polimi.it} \\
  \href{https://orcid.org/0000-0001-8724-1541}{ORCID: 0000-0001-8724-1541} \\
  \And
  Vincenzo Scotti \\
  Karlsruhe Institute of Technology \\
  Karlsruhe, Germany \\
  \texttt{vincenzo.scotti@kit.edu} \\
  \href{https://orcid.org/0000-0002-8765-604X}{ORCID: 0000-0002-8765-604X} \\
  \And
  Matteo Giovanni Rossi \\
  Politecnico di Milano \\
  Milan, Italy \\
  \texttt{matteo.rossi@polimi.it} \\
  \href{https://orcid.org/0000-0002-9193-9560}{ORCID: 0000-0002-9193-9560} \\
}

\begin{document}

\maketitle

\begin{abstract}
Conversational agents are increasingly used as support tools along mental therapeutic pathways with significant societal impacts. 
In particular, empathy is a key non-functional requirement in therapeutic contexts, yet current chatbot development practices provide no systematic means to specify or verify it.
This paper envisions a framework integrating natural language processing and formal verification to deliver empathetic therapy chatbots. 
A Transformer-based model extracts dialogue features, which are then translated into a Stochastic Hybrid Automaton model of dyadic therapy sessions. Empathy-related properties can then be verified through Statistical Model Checking, while strategy synthesis provides guidance for shaping agent behavior.
Preliminary results show that the formal model captures therapy dynamics with good fidelity and that ad-hoc strategies improve the probability of satisfying empathy requirements.
\end{abstract}

\vspace{0.3em}

\section{Introduction}
\label{sec:introduction}

\ac{nlp} tools and techniques are fueling a transformation of the \emph{mental healthcare} domain.
Automated tools and applications for therapy and counselling like \emph{WoeBot} \citep{fitzpatrick-etal-2017-delivering} and \emph{Wysa} \citep{inkster-etal-2018-empathy} are available to anyone through their smart devices. 
These \emph{chatbots} (or \emph{conversational agents}) for therapy offer support and guidance to individuals facing emotional and psychological challenges \citep{kretzschmar-etal-2019-phone,AHMED2021100012}.
In psychotherapy, the psychologist's emotional response to the patient's statements is a key factor for the success of the treatment \cite{eisenberg1987relation}.
Specifically, empathetic behaviour is essential to foster the pro-social attitude of the patient towards the therapist and deliver the therapy correctly \citep{DBLP:journals/tiis/PaivaLBW17}, where we define \emph{empathy} in affective terms as the capability to match another's affective state \cite{feshbach1968empathy,stotland1969exploratory}. 

Therapy chatbots are developed using rule-based systems \citep{kretzschmar-etal-2019-phone}; these approaches facilitate the control of the chatbot's behaviour, but hinder their naturalness and flexibility \citep{jurafsky-martin-2023-speech}.
\emph{Open-domain} chatbots are instead developed using \emph{deep learning}-based techniques and \acp{llm} \citep{scotti-etal-2023-primer}.
\acp{llm} display unprecedented capabilities as chatbots, mastering multiple tasks and languages, but lack a sound control system and are prone to several risks \citep{DBLP:journals/corr/abs-2112-04359,corbo2025toxic}.

Paralinguistic analysis investigates a speaker's \emph{affect}, a manifestation of individual states. Affective computing \citep{picard2000affective} analyzes these states, and empathetic computing extends this by focusing on both cognitive and emotional aspects \citep{DBLP:journals/coling/ZhouGLS20,DBLP:journals/cogsr/Yalcin20}. For conversational agents, empathy promotes human-like responses by enabling appropriate cognitive and emotional reactions \citep{DBLP:journals/coling/ZhouGLS20,DBLP:conf/aaai/LinXWSLSF20}. 
Affective computing techniques, though, are rarely integrated into software engineering processes that ensure correctness, compliance, and traceability. This gap hinders the dependability of chatbots in socio-critical domains.

Therefore, the challenge of engineering empathetic conversational agents emerges \citep{inkster-etal-2018-empathy,DBLP:conf/icassp/ShinXMF20,DBLP:conf/aaai/LinXWSLSF20}.
The societal impact of mental healthcare applications raises the demand for \emph{sound} and \emph{robust} behavior from deep learning-based chatbots \cite{silva2023modeling}. 
In fact, there is a growing conversation on \ac{sleec} requirements and their perspective centrality in the development of autonomous agents \cite{feng2024analyzing}.
From a software engineering perspective, current chatbot development practices, which focus on data and model performance, lack systematic ways to capture, specify, and verify \ac{sleec} requirements. 
Thus, there is a pressing need for SE-oriented solutions that combine modeling, requirements specification, and formal verification to ensure empathetic behavior.

{\vspace{.1cm}
\setlength{\parindent}{0cm}
\sloppypar{\textbf{Contribution.}}
We envision a framework for engineering empathetic therapy chatbots spanning from training to inference. 
The framework combines dialogue analysis, model generation, and verification techniques to synthesize and analyze a \ac{sha} model \cite{DBLP:books/daglib/0030674} of dyadic (\ie involving two subjects) therapy sessions.
Non-functional requirements express desirable properties for the agent's affective and empathetic behavior to be verified.
Strategies are then derived to drive, in addition to dialogue datasets, the agent's behavior through in-context learning.}

The paper reports a preliminary study performed on a prototype of the envisioned framework. 
Through a fine-tuned \emph{Transformer Language Model} (LM) \citep{DBLP:conf/nips/VaswaniSPUJGKP17}, we extract high-level dialogue attributes of conversational phases and speakers' emotional states from field-collected data.
The \ac{sha} model is then automatically generated. The latter incorporates
mined attributes of the dialogue utterances (\eg dialogue act, emotional status) and captures the evolution of the status and behaviour of the speakers participating in the dialogue.
Properties concerning the speakers' behaviour and empathy can then be verified through \ac{smc} \citep{DBLP:reference/mc/2018}. 
Preliminary results show that the \ac{sha} model achieves satisfactory accuracy and highlight trade-offs between different strategies.  

{\vspace{.1cm}
\setlength{\parindent}{0cm}
\sloppypar{\textbf{Related Work.}}
Prior studies on therapy chatbots emphasize opportunities and risks. Haque et al. survey mobile mental-health agents \cite{haque2023overview}, while 
Khawaja et al. highlight the fragility of therapeutic trust \cite{khawaja2023your}, and Seitz \cite{seitz2024artificial} warns against the illusion of ``artificial empathy.'' 
Formal verification approaches have also been explored: conversational flows have been modeled in \uppaal \cite{silva2023modeling}, Brännström et al. detect manipulative dialogues formally \cite{brannstrom2025formal}, and VeriPlan integrates \ac{llm} planning with model checking \cite{lee2025veriplan}. 
Finally, broader reviews stress unresolved ethical and clinical concerns \cite{denecke2021artificial}. 
These works show a clear need for frameworks that link empathetic behaviour with formal verification, a gap our contribution seeks to address.}

{\vspace{.1cm}
\setlength{\parindent}{0cm}
\sloppypar{\textbf{Paper Structure.}}
\sref{fw} presents the framework detailing the \ac{nlp} and formal verification phases;
\Cref{sec:exp} outlines the preliminary evaluation we conducted on our framework;
\Cref{sec:conclusion} concludes the paper and presents future research directions.}

\section{Proposed Framework}
\label{sec:fw}

\begin{figure}[t]
\begin{center}
\includegraphics[width=.8\columnwidth]{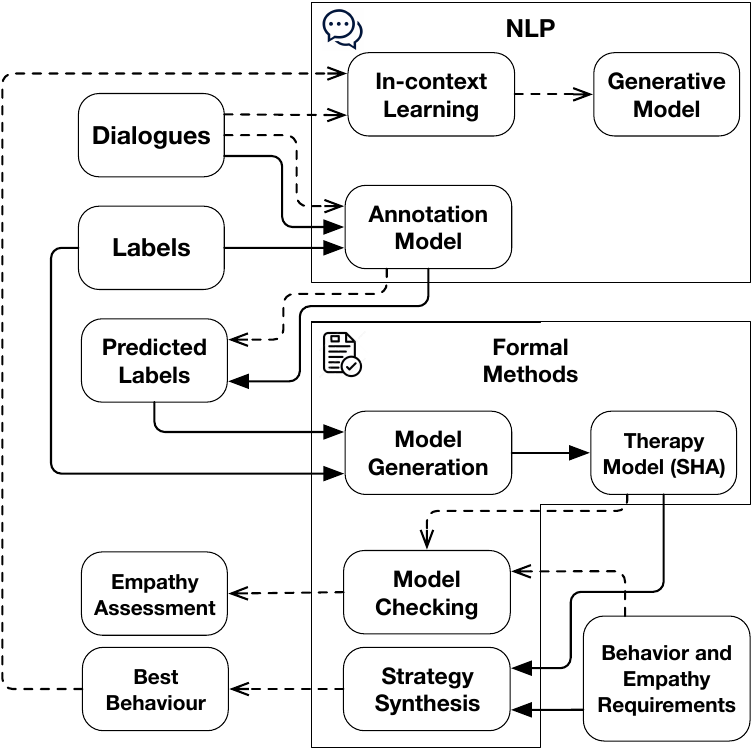}
\caption{Envisioned approach: solid and dashed arrows represent training and inference tasks, respectively.}
\label{fig:workflow}
\end{center}
\end{figure}

We present the envisioned workflow for the framework (see \fref{workflow}) and then  the selected \ac{nlp} and verification techniques.

The proposed framework treats empathy as a non-functional requirement that must be elicited, specified, and verified alongside functional dialogue capabilities. 
Stakeholders such as clinicians, human-computer interaction researchers, and engineers contribute by respectively defining therapeutic goals, mapping them to dialogue features, and formalizing them as verifiable properties.
The workflow structures the lifecycle of a conversational agent: 
\begin{enumerate*}[label=(\roman*)]
    \item requirements are operationalized into dialogue features and formal properties,
    \item a formal model is generated to capture therapy dynamics, 
    \item verification determines whether requirements are satisfied before deployment. 
\end{enumerate*}
By grounding dialogue analysis and verification in explicit engineering steps, the framework enables traceability from stakeholder expectations to verifiable properties on the \ac{sha} model.

At training time, through dialogue analysis techniques, an annotation model processes dialogue datasets and a label catalogue to extract relevant dialogue features (\eg phases, acts, and affective aspects).
The mined attributes are then fed to a formal model generator to create the \ac{sha} model of therapy sessions with a data-driven approach. 
It is then possible to synthesize a strategy for the agent (whose behavior is distilled into the \ac{sha} model) that guarantees behavior and empathy requirements.

At inference time, dialogue features extracted from the patients' statements can be paired with the synthesized strategy to identify the preferable act based on the current context and influence the conversational agent's response through in-context learning. 
Moreover, empathy requirements expressed as logic properties can be verified on the \ac{sha} model through \ac{smc}. Stakeholders are then informed by verification results on the agent's empathy levels. 

\subsection{\acl{nlp} Techniques}
\label{sec:nlp}

\sloppypar{\textbf{Linguistic and Paralinguistic Analysis.}}
Dialogue analysis techniques are necessary to extract the high-level features (i.e., labels) about the dialogue and the speakers from the the \emph{low-level} data (raw text and speech inputs).
\emph{Computational linguistics}, in particular \emph{semantics} and \emph{pragmatics}, offers the tools to understand the meaning and intentions related to the utterances in a dialogue by analysing \emph{what} is being said or written \citep{jurafsky-martin-2023-speech}.
\emph{Computational paralinguistics} highlights information on the \emph{traits} and \emph{states} of the speakers in a dialogue by analysing \emph{how} something is being said or written \cite{schuller2013computational}.

Structural features define a dialogue's formal structure. 
Linguistics (pragmatics) views dialogues as a sequence of phases, with most dialogue structure models employing a finite-state machine approach at varying granularities \citep{VENTOLA1979267,DBLP:conf/sigdial/GilmartinSVCW18}. Speakers take turns, performing dialogue \emph{acts} to advance the conversation. These acts interpret utterances as actions reflecting speaker intents (\eg asking or informing). 
Examples of dialogue act models are the \emph{Dialogue Acts Markup in Several Layers} (DAMSL) notation \citep{allen1997draft}, the \emph{S-Scheme} \citep{DBLP:conf/coling/MezzaCSTR18,DBLP:conf/coling/MezzaW022}, or the hierarchical model proposed in the MEMO \citep{DBLP:conf/kdd/SrivastavaSLA022} corpus of therapy and counselling dialogues.
We consider the following structural aspects, exemplified in \fref{dialogue}: \emph{dialogue phase}, \emph{dialogue acts}, and \emph{utterance duration}.

Monitored features include emotion and empathy communication. Emotion, a temporary individual status analyzed paralinguistically, is modeled via categorical models (\eg Ekman's six basic emotions\cite{ekman1992facial}) or dimensional models.
The latter use a multidimensional continuous representation, where points represent emotions and dimensions represent orthogonal aspects of an individual's emotional state. The circumplex model of affect defines emotional status by \emph{valence} (pleasantness) and \emph{activation} (intensity) \cite{russell1980circumplex}. 

{\vspace{.1cm}
\setlength{\parindent}{0cm}
\sloppypar{\textbf{Pre-trained Language Models}}}
State of the art NLP applies self-supervised learning to train deep neural networks working as probabilistic language models~\cite{radfordimproving,DBLP:conf/naacl/DevlinCLT19}. 
These neural network, based on the Transformer architecture~\cite{DBLP:conf/nips/VaswaniSPUJGKP17}, are pre-trained on large datasets of diverse documents to learn a generative model of text. 
Implicitly, neural networks learn hidden representations that make the model suitable for different tasks involving text analysis.

These models can be used to apply linguistic and paralinguistic analysis in two different ways.
They can be either fine-tuned (i.e., further trained on domain-specific data) to become classifiers and regressors~\cite{radfordimproving,DBLP:conf/naacl/DevlinCLT19} or, especially in the case of the most recent and larger models, they can be used ``as-is'' via in-context learning through prompting~\cite{DBLP:conf/nips/BrownMRSKDNSSAA20}.
The latter case often requires to use models larger than a billion parameters that can be instructed via natural language. 
This allows for a better re-use of the pre-trained models, higher flexibility, and removes the need of having many labelled samples to learn from at the cost of deploying a model that is more computationally demanding and, thus, slower at inference. 

Besides dialogue annotation, these pre-trained models can be used (again via fine-tuning or prompting) to generate response suggestions~\cite{DBLP:journals/corr/abs-1901-08149}.
In fact, the LLMs can be fine-tuned or instructed to generate responses following some high-level specifications (e.g., dialogue act or topic) that can be suggested by some strategy~\cite{DBLP:journals/corr/abs-1901-08149}.

\begin{figure}[t]
    \begin{subfigure}{\columnwidth}
        \centering
        \includegraphics[width=.9\textwidth]{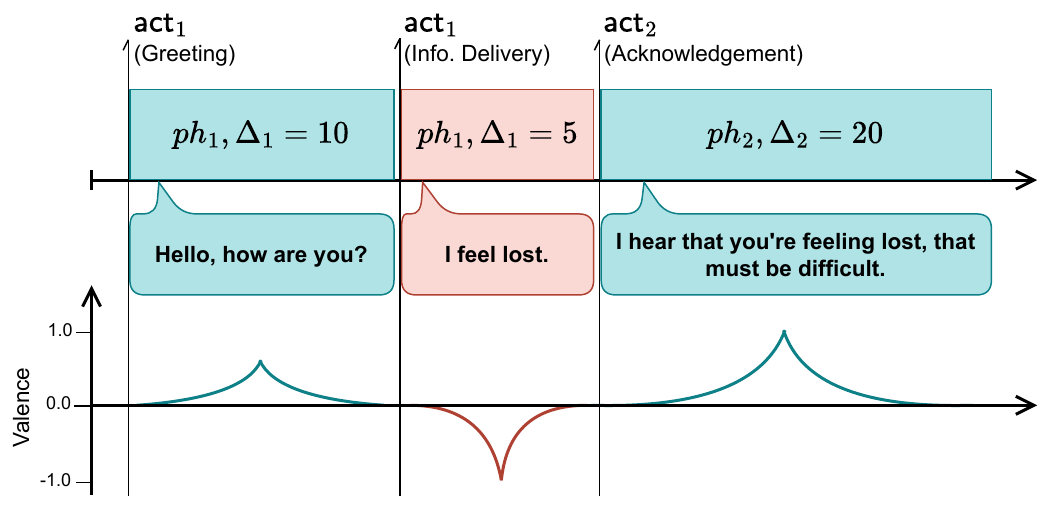}
        \caption{Dialogue features (therapist is in blue, patient is in red).}
        \label{fig:dialogue}
    \end{subfigure}
    \begin{subfigure}{\columnwidth}
        \centering
        \includegraphics[width=.9\textwidth]{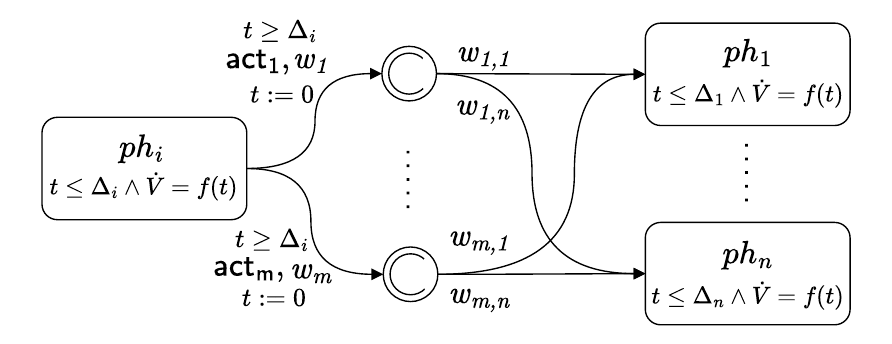}
        \caption{Structure of the \sha{} modeling a speaker.}
        \label{fig:sha}
    \end{subfigure}
\caption{Dialogue model.}
\label{fig:model}
\end{figure}

\subsection{Formal Verification Techniques}
\label{sec:formal}

\sloppypar{\textbf{Formal Model Generation.}}
\label{sec:sha}
In \ac{sha}, \emph{locations} represent system states. 
The hybrid component incorporates real-valued variables with non-linear dynamics governed by \acp{ode} (\emph{flow conditions}) \citep{alur1995algorithmic}. 
\emph{Edges} model discrete transitions, labeled with a triggering \emph{event}, and optionally a guard condition and an update. 
\ac{sha} can form a \emph{network}, synchronizing through events and performing simultaneous transitions. 
Edges can be probabilistic, with weights indicating the system's bias towards choices.

Our framework models a dyadic therapy session with two \ac{sha}, one for each speaker (patient and therapist), and is sufficiently general to capture both human and chatbot therapists. 
\fref{sha} illustrates the snippet of a speaker's \ac{sha}, assuming $m$ dialogue acts and $n$ dialogue phases. 
Each location corresponds to a dialogue phase.
Dialogue acts are event labels. Performing a dialogue act triggers a two-edge transition between phases. First, from phase $ph_i$, an edge labeled $\mathsf{act}_k$, with $k\in[1, m]$, leads to a \emph{committed} location. In committed locations, time does not elapse; an outgoing edge immediately fires to a destination phase $ph_j$, with $j\in[1, n]$. 

Edges have probability weights: $w_j$ captures the probability of performing $\mathsf{act}_j$ while in $ph_i$, whereas $w_{j, i}$ captures the probability of reaching phase $ph_i$ after performing $\mathsf{act}_j$.
Decoupling these edges preserves the network's input determinism (no location has multiple outgoing edges with the same event), a pre-condition for \ac{smc}. 
Locations and probability weights are extracted from dialogue datasets: weights for $\mathsf{act}_j$ edges are relative frequencies of dialogue acts given a phase; for edges from committed locations to $ph_i$, they are relative frequencies of next phases given a current phase and dialogue act.

Locations also determine the duration of a dialogue phase and the evolution of the speakers' emotional status. 
Clock $t$ models time in $ph_i$. Invariant ${t\leq\Delta_i}$ and guard ${t\geq\Delta_i}$ on outgoing edges force a dialogue act after exactly $\Delta_i$ time units. 
Emotional status uses a dimensional notation with three dimensions as per the circumplex model. 
The dynamics of valence (variable $V$) in a phase follows flow condition $f(t)$ (see \fref{dialogue} for an example). 
In the envisioned framework, parameters $\Delta_i$ and function $f(t)$ are estimated based on field-collected data and analytic models from the literature \citep{fan2019minute}.

{\vspace{.1cm}
\setlength{\parindent}{0cm}
\sloppypar{\textbf{Statistical Model Checking.}}
Given \sha{} network $M$, through Monte Carlo simulations, \ac{smc} calculates the \emph{probability} of a property $\psi$ in \ac{mitl} to hold within a time bound $\tau$ (corresponding to expression $\mathbb{P}_M(\diamond_{\leq\tau}\ \psi)$, where $\diamond$ is the ``eventually'' operator) \cite{david2015uppaal}.}
In our work, properties express behavior and empathy requirements, such as
the patient's valence always belonging to an acceptable range, which corresponds to expression
$\mathbb{P}_M(\square_{\leq\tau} V\in[\mathsf{V_{min}}, \mathsf{V_{max}}])$
(where $\square$ is the ``always'' operator). 

{\vspace{.1cm}
\setlength{\parindent}{0cm}
\sloppypar{\textbf{Strategy Synthesis.}}
Given dialogue data and the \ac{sha} model, strategies are synthesized to resolve the model’s probabilistic choices. 
Concretely, this involves selecting dialogue acts at each conversational state according to different criteria, such as a purely stochastic baseline, empirical frequencies extracted from data, or optimization heuristics that maximize the patient's valence. 
By embedding these policies into the formal model, we obtain strategized \ac{sha} networks whose executions reflect distinct behavioral patterns of the agent, enabling comparative verification of empathy-related requirements.
}

\section{Preliminary Evaluation}
\label{sec:exp}

This section presents the initial experiments we conducted (\Cref{sec:experiments}) and the early findings (\Cref{sec:results}).\footnote{Replication package available at: \url{https://zenodo.org/records/17226928}.}

\subsection{Experimental Setup}
\label{sec:experiments}

{
\setlength{\parindent}{0cm}
\sloppypar{\textbf{Selected Datasets.}}
We used the MEMO corpus of counselling and therapy conversations \citep{DBLP:conf/kdd/SrivastavaSLA022} as the main reference for our work, since it contains domain-specific dialogues annotated with some of the desired features---specifically, the dialogue phase (referred to as \emph{counselling components}) and the dialogue acts---using a hierarchical model.
Additionally, we used the IEMOCAP corpus \citep{DBLP:journals/lre/BussoBLKMKCLN08} to extract statistics about the duration of the utterances in dialogue.
IEMOCAP is a multimodal dialogue corpus that also offers speech recordings of the individual utterances in the dialogues.
We used the speech recordings to extract the duration of the utterances.}

The IEMOCAP data set \citep{DBLP:journals/lre/BussoBLKMKCLN08} was created to help build emotion recognition systems and contains dialogues labelled with the emotions of the speakers using both categorical and dimensional notations.
The EPITOME data set \citep{DBLP:conf/emnlp/SharmaMAA20} contains message-response pairs from peer-to-peer support conversations extracted from an online platform.
The pairs are annotated with the strength of three communication mechanisms describing the empathy of the response: \emph{emotional reaction}, \emph{interpretation}, and \emph{exploration}.

{\vspace{.1cm}
\setlength{\parindent}{0cm}
\sloppypar{\textbf{Formal Modeling and Verification Tools.}}
For the formal modelling phase, we employ \uppaalsmc{} \citep{david2015uppaal}, an integrated environment for modelling, validation, and verification of real-time systems, which supports the \ac{sha} formalism. 
The generated formal model conforms to the modeling approach described in \sref{formal}.
Through \uppaalsmc, it is then possible to calculate the probability of properties holding for the \ac{sha} network under analysis, where properties are expressed in the \ac{mitl} notation.
}

{\vspace{.1cm}
\setlength{\parindent}{0cm}
\sloppypar{\textbf{Settings.}} 
With the selected datasets, we conducted two distinct experiments. 
First we synthesised a formal model starting from the dialogue stages and dialogue acts labelled in the MEMO corpus, which we augmented with information about patient valence to include the emotional information using a model trained on IEMOCAP.
Secondly, starting from the synthesised formal model, we built examples of possible strategies aimed at suggesting the dialogue act to the therapist and we evaluated them on the MEMO data.}

We evaluated the formal model of the therapy dialogue under two aspects: transition probability and expected valence. 
For both aspects, we compared model predictions with test data coming from the MEMO corpus.
Using \uppaalsmc{}, we compared the expected and observed transition probabilities on the test dialogues and we compared the expected and observed valence values.

\begin{figure}[t]
\centering
\includegraphics[width=\columnwidth]{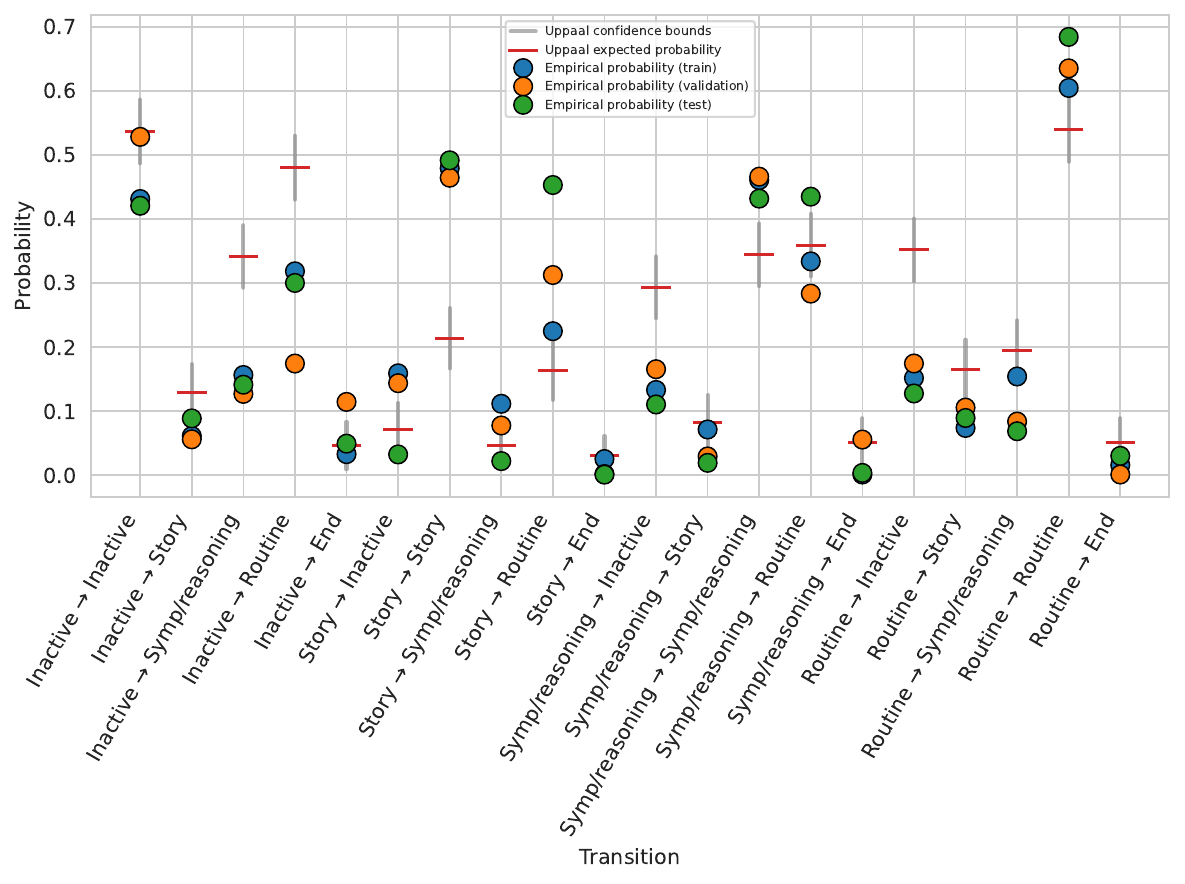}
\caption{Transitions compliance of observations from test data with UPPAAL estimated confidence bounds}
\label{fig:modelcomplianceresults}
\end{figure}

Concerning the policy evaluation, we compared three different strategies:
\begin{enumerate*}[label=(\arabic*)]
 \item \emph{Random} (baseline selecting the dialogue act randomly),
 \item \emph{Data-driven} (model selecting dialogue acts following the empirical frequencies observed in the training data), and
 \item \emph{High Valence} (for each stage, select the dialogue act that yields the highest improvement in patient valence, following the idea of \emph{sentiment lookahead}~\cite{DBLP:conf/icassp/ShinXMF20}).
\end{enumerate*}
We compared the three strategies computing the probability of three empathy requirements under each of them:
\begin{enumerate*}[label=(REQ\arabic*)]
    \item sympathetic reasoning is always reachable, corresponding to \ac{mitl} formula  $\mathbb{P}_M(\lozenge_{\leq\tau}\ \mathrm{therapist.symp\_resoning})$;
    \item the session ends in a sufficiently positive state (formula $\mathbb{P}_M(\lozenge_{\leq\tau}\ \mathrm{patient.end}\land\mathrm{patient.}V >\mathsf{V_{min}})$);
    \item a sufficiently positive state is sustained for at least $\mathsf{T_{min}}$ time units (formula $\mathbb{P}_M(\lozenge_{\leq\tau}\ \mathrm{patient.}V >\mathsf{V_{min}} \land t\geq\mathsf{T_{min}})$).
\end{enumerate*}

\subsection{Results}
\label{sec:results}

{
\setlength{\parindent}{0cm}
\sloppypar{\textbf{Dialogue Model.}}
The comparison between model-generated and empirical transition probabilities from the data set showed that, despite its simplicity, the formal model captures the main patterns of therapeutic dialogue with good fidelity. 
\Cref{fig:modelcomplianceresults} shows the transition probabilities for each stage, highlighting where predicted transitions between stages fell within the observed probability ranges. 

{\vspace{.1cm}
\setlength{\parindent}{0cm}
\sloppypar{\textbf{Strategy.}}
The off-policy evaluation of the dialogue strategies further confirmed the usefulness of the model. 
As per \tref{strategyresults}, the strategy that maximizes patient sentiment satisfies all requirements with higher probability then the random and the data-driven strategies.} 

The data-driven and random strategies produced results very similar to one another, reflecting the average neutral character of the dataset. 
This result suggests that only optimising the cumulative variation of valence,
despite representing a valid approach for empathy~\cite{DBLP:conf/icassp/ShinXMF20}, may not be the only objective followed by the therapist or the counsellor.
Thus, we may need to reconsider the objective function for the policy. 

\begin{table}[t]
\centering
\caption{\ac{smc} Results by Requirement and Strategy}
\label{tab:strategyresults}
\resizebox{.8\columnwidth}{!}{
\begin{tabular}{lcccc}
\toprule
{\multirow{2}{*}{\textbf{REQ}}} & \multicolumn{3}{c}{\textbf{Strategy}} \\[2pt] \cline{2-4}
& \textit{High Valence} & \textit{Random} & \textit{Data-driven} \\
\midrule
\textbf{1} & [0.7410, 0.8350] & [0.6696, 0.7659] & [0.6716, 0.7678] \\
\textbf{2} & [0.1994, 0.2947] & [0.1692, 0.2636] & [0.1007, 0.1915] \\
\textbf{3} & [0.9390, 0.9997] & [0.8641, 0.9500] & [0.9112, 0.9872] \\
\bottomrule
\end{tabular}}
\end{table}

\section{Future Plans}
\label{sec:conclusion}

This paper presents our preliminary work toward applying formal verification to behaviour and empathy assessment in therapy dialogues. 
The preliminary pipeline exploits NLP tools (namely fine-tuned Transformer LMs) to extract the high-level features necessary for a formal model (based on \sha) to run its verification.
From this starting point, we envision the development of a tool to correct the current behaviour of a therapy or counselling chatbot towards a more empathetic one. 
From a technical perspective, we plan to improve the empathy evaluation and empathetic policy synthesis, defining a better objective function based on models extracted from EPITOME.
Additionally, we plan to complete our framework by implementing a module that generates suggested responses based on the empathetic strategy predictions.

\bibliographystyle{plain}

\end{document}